\pdfoutput=1
\documentclass[11pt]{article}

\usepackage[]{NLPAICS2024}

\usepackage{times}
\usepackage{latexsym}

\usepackage[T1]{fontenc}
\usepackage[utf8]{inputenc}

\usepackage{microtype}

\usepackage{inconsolata}
\usepackage{graphicx}

\title{Variation between credible and non-credible news across topics}

\author{Emilie Marie Carreau Francis\\
  Språkbanken Text \\
  University of Gothenburg, Sweden\\
Box 200, SE 405 30 Gothenburg\\
  \texttt{emilie.francis@gu.se} %\\\And
}

\begin{document}
\maketitle
\begin{abstract}
`Fake News' continues to undermine trust in modern journalism and politics. Despite continued efforts to study fake news, results have been conflicting. Previous attempts to analyse and combat fake news have largely focused on distinguishing fake news from truth, or differentiating between its various sub-types (such as propaganda, satire, misinformation, etc.) This paper conducts a linguistic and stylistic analysis of fake news, focusing on variation between various news topics. It builds on related work identifying features from discourse and linguistics in deception detection by analysing five distinct news topics: Economy, Entertainment, Health, Science, and Sports. The results emphasize that linguistic features vary between credible and deceptive news in each domain and highlight the importance of adapting classification tasks to accommodate variety-based stylistic and linguistic differences in order to achieve better real-world performance.
\end{abstract}

\section{Introduction}
\label{sec:intro}
The term `Fake News' catapulted to popularity around the 2016 U.S. Presidential election and has continued to cast a shadow of mistrust over journalism and politics \citep{Ram2023,Volz2023}. Global trust in social media as a news source remains low and has been on a decline for the past decade \citep{Edelman2024}. Despite this, many still turn to social media as a means to stay informed. Half of the U.S. adult population report getting their news from social media at least some of the time \citep{Wang2024, Matsa2023}. However, most users express concerns about quality, accuracy, and bias \citep{Wang2024}.

The effort to combat the spread and influence of `fake' or `non-credible' news has been reflected in the large body of academic research on fake news detection and analysis. However, there has been little large scale practical implementation of this research. In part, this can be attributed to conflicting observations in the literature. This paper takes a variety specific approach to non-credible news analysis by investigating linguistic and stylistic differences for five common news topics: economy, entertainment, health, science/technology, and sports. 

\subsection{Contributions}
Previous approaches to fake news analysis and detection have taken either a broad view of news, by disregarding or combining news categories, or focusing only on hard news. The inclusion of linguistic and stylistic features in automatic classification is promising, but results remain lacklustre. Models may be sensitive to genre/domain attributed differences and could benefit from more targeted classification approaches. This research investigates differences between credible and non-credible news across a variety of contexts to provide support for this assumption, in addition to the introduction of a novel topic-based `fake news' dataset. The following questions will be addressed:

\begin{enumerate}
	\item What are the stylistic differences of non-credible and credible news for each topic?
	\item What differences (if any) are observed across topics?
\end{enumerate}

The goal of these questions is to determine the generalisability of stylistic based fake news detection and identify features which may be used in classification models. This research will also consider how such cues agree or contradict previous literature on deceptive and persuasive language in journalism and politics.

\section{Related work}
\label{sec:relwork}
\paragraph{Deceptive and persuasive language:} In political communication, advocates attempt to manipulate the public in many ways. Arguments can be classified into four types depending on whether it is pro, con, easy, or hard to comprehend \citep{Cobb1997}. To assess the persuasive power of each argument type, \citet{Cobb1997} studied opinions on the North American Free Trade Agreement (NAFTA) and healthcare at three points in time. Oppositional arguments held more weight, and for NAFTA the effect was stronger for hard arguments. However, easy arguments were more persuasive for healthcare. In policy proposal, \citet{Lau1991} observed that persuasion can be influenced by the formulation and presentation of interpretations. An argument is more persuasive, regardless of a voter's political beliefs, if one can control the environment to allow for only one interpretation.

`Control over the narrative' is often a factor in identifying propaganda. Journalism uses four factors to distinguish persuasion from propaganda: volition, transparency, manipulation, and the shielding of listeners from opposing facts \citep{Bard2017}. Propagandists exploit audience beliefs and values to promote self-interest, attempt to block opposing arguments from reaching the audience, and often hide the true intent of their message. \citet{Simpson1992} argues that lying involves three levels: deception regarding a state of affairs, regarding one's beliefs, and regarding the sincerity of one's presentation as believing. The third level distinguishes simply being untruthful from legitimate deception, as to be untruthful is not necessarily to lie. This third level differentiates satire from simple misinformation, but can also be used to occlude the intents of many fake news creators who claim there is no reason for readers to believe their content is sincere. 
 
Studies on lying have revealed that certain cues can be used to indicate deception. It has been shown that the psychological burden of lying, whether due to guilt or the challenge of remembering the lie, may cause liars to avoid language that takes ownership of the statement or portray certainty \citep{Newman2003,Wawer2023,Dzindolet2005}. \citet{Newman2003} found that, in addition to using more words that elicit negative emotion, liars distance themselves from claims by using fewer first and third person pronouns. In a study of true and false statements in Polish and English, \citet{Wawer2023} also observed that lying in English triggered an increase of words with negative tone, as well as a general increase in negation.

\paragraph{Deception in news text:} While there have been several studies investigating linguistic and stylistic features in news text, results are often contradictory. \citet{Potthast2018} used style analysis, including readability scores and dictionary features, to distinguish hyper-partisan from mainstream news. It was found that left and right-wing news share more stylistic similarities with each other than with mainstream. Writing style on its own was discovered to be sufficient for distinguishing hyper-partisan news articles from more balanced news. \citet{Mahyoob2020} found that proper nouns and passive voice is more frequent in credible news, while non-credible news uses more superlatives. While identifying linguistic features to use in automatic classification, \citet{Kasseropoulos2021} noted that fake articles are shorter in length and use fewer technical words, quotes, punctuation, and have more lexical redundancy. They also use simpler language with shorter words, as well as more personal pronouns and adverbs. 

In satire, surface level features such as sentence length and average word frequency, in addition to semantic features and causal connectives vital to text comprehension are considered predictors \citep{Levi2019}. Disinformation is also prone to grammatical and orthographical mistakes, erratic punctuation, and idiosyncratic typography \citep{Silva2022}. Another study investigated variation between credible and deceptive hard news and found that credible news was more informationally dense, while deceptive news was more narrative \citep{Francis2018}. It was also observed that credible news used more adjectives, intensifiers, and clausal coordination. 

\citet{Addawood2019} used interpersonal deception theory (IDT) and reality monitoring (RM) to analyse the language used by Russian trolls during 2016 U.S. Presidential election. They identified 49 linguistic cues used by deceivers which indicated uncertainty, including hedges, modal verbs, auxiliary verbs, and expressions of possibility. It was also observed that deceivers attempted to distance themselves from the lies by using less self-reference in the form of pronouns. In research on climate change news and editing of international news, a trend of misreporting illustrated by errors such as overstatement, misquotation, misattribution, and over-assertion, was revealed \citep{Bell1991}.

It also appears that differences between fake and credible news vary across languages and dialects. In a study of English and Portuguese fake news texts, \citet{Silva2022} noticed that variation between fake news and mainstream media differ depending on the corpus. The longest words are used by fake news in the English corpus, whereas the longest words are used by mainstream media in Portuguese. In a study of false statements in Brazilian Portuguese news, \citet{Vargas2021} used word and sentence level analysis to discover that true statements used more nouns and verbs, while false statements were found to contradict previous literature in their pronoun usage.

In depth discourse and linguistic analysis of deceptive text has also revealed some interesting trends throughout the various forms of fake news. Using van Leeuwen’s discourse model of legitimation and de-legitimation, \citet{Igwebuike2021} analysed the legitimation strategies used for justification of fake news posts on Nigerian WhatsApp, Facebook, and Twitter. Findings revealed that creators convey messages to readers and validate disinformation through appeals to authority, emotion, moralisation, and rationalisation. 

\paragraph{Deception detection:}  \citet{Verma2021} used linguistic features to classify the veracity of news content by organizing the features into sets and merging them with word embeddings. The 20 most salient features were selected and applied to a voting classifier. \citet{Burgoon2003} employed 16 linguistic features categorized into four classes in a decision tree algorithm, achieving an accuracy of 60.72\%. \citet{Vicario2018} used a variety of features from text (e.g. number words, sentences, and characters), along with user and message specific features to identify hoaxes and fake news on social media with various machine learning models.

Introducing the small novel UNBiased dataset, \citet{Gravanis2019} tested 57 linguistic features embedded with word-to-vector embedding in several popular ML classifiers for deception. \citet{Kasseropoulos2021} identified an optimal set of 23 features out of 87 which performed well with CNN and LSTM classifiers. LUX (Language Under eXamination), is a text classifier that makes use of linguistic analysis to infer the likelihood of an input being fake-news \citep{Azevedo2021}. Linguistic metrics were included as model features to improve classification performance in identifying fake news. 

Many other approaches to automatic deception detection have made use of shallow text features and semantics in their models with reasonable success in the task at hand \citep{Bharadwaj2019,Kurasinski2020}. \citet{Kuzmin2020} used models trained on bag-of-n-grams and bag-of-RST (Rhetorical Structure Theory) features to detect satire, real, and fake news in Russian and discovered that unigrams were the most important feature for detection. In a survey of supervised learning approaches to deception detection with discourse and structural features, the results of such approaches were mixed, but showed promise \citep{Vargas2022}.

The results from previous literature show a conflicting landscape of deceptive language in writing and news text. Likely due to this inconsistency, automatic deception detection methods which have utilised linguistic features exclusively typically achieve lukewarm performance. One of the limiting factors of the previous approaches is that analysis has been broadly based on hard news and politics with little investigation into other news topics. The research presented in this paper adds to the existing body of literature by including under-represented news topics, such as entertainment and sports. The results of this paper will reference the previous findings detailed above, identifying similarities and differences. The following sections will introduce the data utilised in this study and the methodology through which they have been analysed.

\section{Data} 
\label{sec:data}
\begin{table}
	\centering
	\begin{tabular}{ll}
		\hline
		\textbf{News Type}             & \textbf{Total Articles} \\ \hline
		\textbf{Economy}               & 15,672                  \\
		\textbf{Entertainment}         & 5,000                   \\
		\textbf{Health}                & 5,258                   \\
		\textbf{Science \& Technology} & 8,400                   \\
		\textbf{Sports}                & 6,842                   \\ \hline
	\end{tabular}
	\caption{\label{data} Total articles per topic. The number of articles per label is equal to the total divided by two.}
\end{table}

A novel dataset was created in order to ensure a balanced sample size with consistent topic labelling. Texts are limited to English news from the United States and Canada during the period of 2011 to 2018. Article annotation was carried out automatically as data was collected. Articles were labelled `non-credible' or `credible' based on publisher intent, similar to the approach taken by \citet{Lazer2018}. Articles from publishers whose mission is perceived as providing accurate information with high reliability are categorized as `credible', whereas articles from publishers who intentionally produce fabricated stories or have mixed/lower factuality ratings are categorized as `non-credible'. 

Labels are determined based on bias and reliability scores provided by Media Bias-Fact Check,\footnote{\url{https://mediabiasfactcheck.com/}} AllSides,\footnote{\url{https://www.allsides.com/}} and Ad Fontes Media.\footnote{\url{https://adfontesmedia.com/}} Bias ratings are determined using a numerical scale, based on various factors (including political leaning, factuality, spin/framing, and several types of bias), averaged from a survey of articles from the outlet. While these companies have slightly different approaches to rating, all employ a panel based system where a selection of articles and headlines from an outlet is reviewed regularly by a balanced panel of raters who have self reported their political biases. As these organisations are private companies, specific guidelines are not publicly available. However, assessment criteria are described in detail on the respective websites.

The labels `credible' and `non-credible' were chosen based on the definition of `credibility' as something trustworthy or worthy of belief. This label covers the range of deceptive topics in the analysis, including satire. Despite the primary intent of satire being entertainment, it is still considered non-credible due its potential to mislead readers and its lack of trustworthiness as a source of information. Furthermore, while automatic classification between hard news and satire has been somewhat successful \citep{Horne2017, Rubin2015}, it is often challenging to distinguish satire from other forms of deceptive news \citep{Rashkin2017}.

Sources for the credible corpus are Reuters, the New York Times, Global News, Business Insider, CBC, and the New Yorker. Non-credible news sources include the Beaverton, Breitbart, Global Research, If You Only News, Your Newswire, MadWorld News, and Liberty Writers. There is a total of 41,172 articles in the combined non-credible and credible news corpus, with a 50:50 split of deceptive and credible news for each topic. Table \ref{data} shows the number of articles for each topic, where the number of articles per label is equal to the topic total divided by two. Smaller numbers for topics such as entertainment and health can be attributed to lower publication rates for those topics in general, especially compared to more hard news like economy. This is compounded for non-credible news which typically has a lower overall publication rate compared to credible news outlets.

\section{Methodology}
\label{sec:meth}
\begin{table*}[]
	\setlength\tabcolsep{0pt}
	\centering\small
	\begin{tabular}{lcccc} 
		\hline
		& \textbf{High Score (H)} &                            & \textbf{Low Score (L)} &                             \\ 
		\hline
		\textbf{D1: Involved vs. Informational}            & Involved                & verbs, pronouns            & Informational          & nouns, adjectives           \\
		\textbf{D2: Narrative vs. Non-Narrative}           & Narrative               & past tense, third person   & Non-narrative          & synthetic negation          \\
		\textbf{D3: Context-Independent vs. Dependent}     & Independent             & nominalisations            & Dependent              & adverbs, pied-piping        \\
		\textbf{D4: Overt Expression of Persuasion}        & Explicit                & modal adverbs              & Absent                 & suasive verbs, infinitives  \\
		\textbf{D5: Abstract vs. Non-Abstract Information} & Abstract                & passive clauses, conjuncts & Non-abstract           & agentless passives          \\
		\hline
	\end{tabular}
	\caption{Five of the six dimensions used in this paper. The `H' and `L' tags represent the text-type associated with high or low scores for the dimension, including characteristic high frequency features for the type.}
	\label{dimension}
\end{table*}
\subsection{Multi-dimensional analysis (MDA)}
MDA is a means of measuring textual variation in text types based on a collection of linguistic features. Six dimensions, each associated with underlying communicative functions, were established to group texts based on similarity of composition \citep{Biber1988}. A text is analysed by tagging linguistic features and calculating a factor score, which is used to represent groupings of linguistic variables observed to have high co-occurrence. Features with a factor magnitude of 1.95 or greater are considered significant to the corpus. 

Dimension scores, calculated from the aforementioned factor scores, determine to which text-type a piece of text is most similar in style. Dimensions one through five are described in Table \ref{dimension}. Dimensions one, two, and four are fairly straightforward, but dimensions three and five may be unclear without further explanation. For dimension three, low scores indicate context dependence and are typical of texts like sports broadcast, whereas context independence is a feature of academic writing. High scores on dimension five indicate that a text presents information in a technical and abstract manner, such as scientific discourse. Dimension six, used to measure informational texts produced under time constraints, is not relevant for this task and has been omitted.

\subsection{Multi-dimensional analysis tagger (MAT) implementation}
Tagging and score calculation were performed with version 1.3.1 of MAT \citep{Nini2019}. MAT is based on the Stanford Part-Of-Speech Tagger and designed to replicate the tagger used in \citet{Biber1988} for multi-dimensional functional analysis of English texts. MAT generates a grammar annotated version of the corpus, in addition to statistics for text-type and genre analysis. \citet{Nini2019} utilises z-scores instead of factor scores, which serve the same function. New tags, such as \textit{indefinite pronoun}, \textit{quantifier}, and \textit{quantifier pronoun}, have been introduced to expand on the original set of features. 

\citet{Nini2019} asserts that MAT provides a good replication of Biber's analysis and has achieved an accuracy of 90\% in similar studies \citep{Grieve2023}. Only the first 400 tokens of an article are used in the analysis, as was the standard used in \citet{Biber1988}. This number may be adjusted, but was determined sufficient to cover the majority of content in the average article. Comparison between credible and deceptive news are discussed using effect size with Cohen's $d$ and Pearson's $r$, assuming the standard guidelines.\footnote{Pearson’s $r$ = .10, .30, and .50, and Cohen’s $d$ = 0.20, 0.50, and 0.80 as small, medium, and large, respectively}

\section{Analysis}
\subsection{Dimension scores} 
\label{dsec}
\begin{figure*}[!ht]
	\begin{center}
		\includegraphics[scale=0.35]{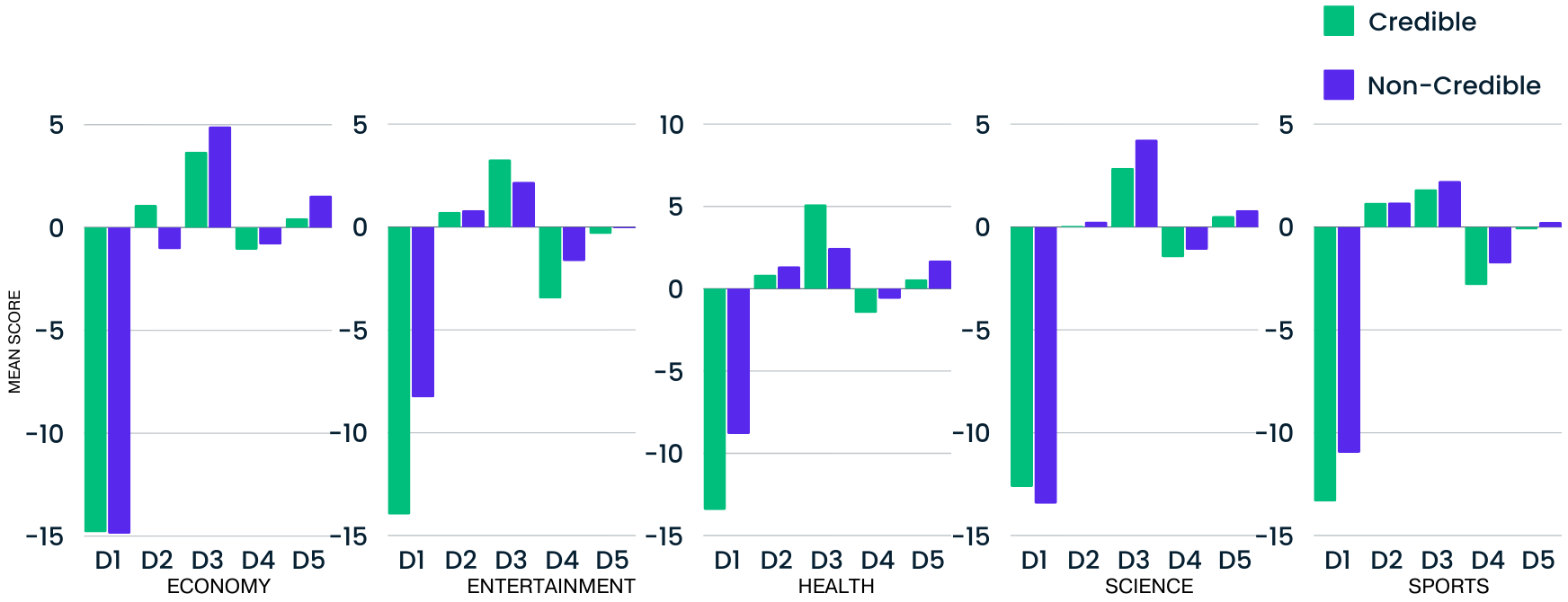} 
		\caption{The mean dimension scores for each news type organized by topic. Non-credible and credible news, with the exception of Economy, scored similarly to the text-type `General Narrative Exposition' defined by \citet{Biber1988}. This is expected, as `General Narrative Exposition' canonically contains news discourse, among other types.}
		\label{dmean}
	\end{center}
\end{figure*}
The corpus with the highest average difference across all dimensions was health news, with economy and entertainment news in a relatively close second. D1 shows the most variation across topics, with the exception of economy news where there is almost no difference. The credible corpus displayed a consistently low D1 score, which indicates that information density is a trait of credible news. While scores for D5 were fairly low for all news topics, non-credible news generally scored higher on D5 compared to credible news. This suggests that non-credible news typically expresses information in a more abstract/technical manner compared to credible news. While this observation appears to contradict research on readability and complexity, it is consistent with Cobb's observations on the persuasive power of hard arguments and previous research on political news discourse \citep{Kasseropoulos2021, Wawer2023, Cobb1997, Francis2018}. Figure (\ref{dmean}) shows the mean score for each dimension by topic and numbers in parentheses below report Cohen's $d$.

\paragraph{Economy:}
\label{d.ec}
The most noticeable difference between deceptive and credible news is in D2. This was also the dimension which showed the greatest effect size (0.87) when comparing means. This suggests that credible economy news language is more narrative while non-credible news is non-narrative. D5 was the other dimension which showed a medium-small (0.45) difference between the corpora, which indicates that non-credible economy news is more formal and technical. There was also a small effect (0.42) in D3, suggesting that non-credible news is moderately more context independent (as the case is with academic prose).

\paragraph{Entertainment:}
\label{d.ent}
Credible entertainment news received a noticeably more negative mean score for D1 compared to non-credible news (Fig.\ref{dmean}). The effect on this dimension is medium-large (0.75), indicating that information in credible entertainment news text is more dense. In contrast to economy news, credible entertainment news demonstrated more context independence than deceptive news. Entertainment also saw a medium-small effect size on D4 (0.45), revealing a difference in expression of persuasion between non-credible and credible news. Overall, credible news employs mildly less persuasion and is somewhat less dependent on context compared to non-credible.

\paragraph{Health:}
\label{d.he}
Health news had the biggest difference between the credible and non-credible corpora altogether, especially between D1 and D3 which showed a medium-large effect (0.69 and 0.73 respectively). This suggests that information in credible health news is more dense and less dependent on context, similar to academic texts. Once again, non-credible news had a higher mean score on D5 compared to credible news, indicating more technical and formal writing. Unlike the other news topics, a medium-small effect was also observed on D2 and D4 (0.41 and 0.42 respectively). Credible health news is mildly less narrative and exhibits less persuasion. The more abstract style taken in non-credible health news may be explained by Cobb's findings on hard arguments.

\paragraph{Science:} 
\label{d.sci}
Differences between credible and non-credible science news were minimal, with most dimensions demonstrating only a small effect. D3 displayed the strongest difference between corpora within this topic, a medium-small effect (0.43). While both received a positive mean score, indicating context independence, the score received by the non-credible corpus was notably higher.

\paragraph{Sports:}
\label{d.spo}
Differences between credible and non-credible sports news were mostly non-existent, but D1 and D4 had a small effect on the corpus (0.34 and 0.30). Although both credible and non-credible news scored low on D1, the mean for credible sports news was noticeably lower. This suggests that credible sports news may contain more information compared to non-credible. A small effect was also observed on D4, indicating that author point of view may be slightly more present in credible sports news. This may be due to the higher likelihood that credible sources attract professional sports writers who offer expert opinions on sporting outcomes, whereas deceptive news writers are more likely to be amateurs. Additionally, since the topic of sports is generally accessible to a wider audience, it is more challenging to present alternative interpretations due to its familiarity \citep{Lau1991}.

\subsection{Linguistic features}
\label{lingsec}
As there are 66 linguistic variables included by MAT, it is unfeasible to discuss all for each news topic. Only features with an unexpected relationship or large difference will be given specific attention. Some strong correlations among linguistic features are what one would expect based on grammar (e.g., a strong negative correlation between adverbs and nouns), so they have been omitted. Nouns and nominalisations were the most frequent across all topics for both corpora, which is expected of general narrative exposition.

\paragraph{Economy:}
\label{ling.ec}
\begin{table}[]
	\resizebox{\columnwidth}{!}{
	\begin{tabular}{lccc}
		\hline
		\multicolumn{1}{c}{\textbf{Feature}} & \textbf{Cohen's $d$} & \textbf{Credible Mean} & \textbf{Non-Credible Mean} \\ \hline
		Conjuncts (CONJ)                     & 0.66               & 0.09                   & 0.91                    \\
		Adjectives (JJ)                      & 0.71               & 0.29                   & 0.93                    \\
		Public Verbs (PUBV)                  & 1.20               & 1.24                   & -0.13                   \\
		Subordinator `that’ deletion (THATD) & 0.77               & -0.17                  & -0.34                   \\
		Past Tense Verbs (VBD)               & 1.00               & 0.03                   & -0.48                   \\ \hline
	\end{tabular}}
		\caption{Features which have a notable effect in the \textbf{Economy} news type. PUBV verbs are identified by \citet{Quirk1985} as those which indicate speech acts. THATD is added when the subordinator is missing from a subordinate clause preceded by a public, private, or suasive verb.}
		\label{econ.ling}
\end{table}
Differences between linguistic features were the most striking in economy news (Table \ref{econ.ling}). The frequency of public verbs was considerably higher in credible economy news compared to non-credible news, and demonstrated the strongest effect in the entire study. Public verbs, such as `say' or `claim', are a major stylistic difference between credible and deceptive economy news. Past tense verbs were also far more frequent in credible news. Subordinator `that' deletion demonstrated a medium effect, which is likely due to the higher frequency of public verbs in the credible news corpus.

In non-credible economy news, there was a considerably greater frequency of conjuncts and adjectives. This observation opposes the findings of \citet{Newman2003}, but is consistent with the opposing observations of \citet{Addawood2019}. A moderate positive relationship was revealed between second person pronouns and conditionals (0.46) and third person pronouns and past tense verbs (0.43) in credible news. In non-credible news, adjectives had the largest effect on average word length (0.53), which is similar to previous observations on hard news \citep{Francis2018}. There is also a mild positive correlation between private verbs and \textit{wh}-clauses (0.31). 

\paragraph{Entertainment:} 
\label{ling.ent}
\begin{table}[]
	\resizebox{\columnwidth}{!}{
	\begin{tabular}{lccc}
			\hline
			\multicolumn{1}{c}{\textbf{Feature}} & \textbf{Cohen's $d$} & \textbf{Credible Mean} & \textbf{Non-Credible Mean} \\ \hline
			Prepositional Phrases (PIN)          & 0.49               & -0.45                   & -0.81                    \\
			Type-Token Ratio (TTR)               & 0.49               & 0.00                   & -0.97                   \\
			Second Person Pronouns (SPP2)        & 0.38               & -0.46                  & -0.26                   \\
			Demonstrative Pronouns (DEMP)        & 0.37               & -0.44                  & -0.17                   \\ \hline
	\end{tabular}}
	\caption{The features with a notable effect in \textbf{Entertainment} news. The preposition `to' is distinguished from the infinitive marker `to' by MAT, receiving the PIN tag. TTR is a measurement of the number of types within the first 400 tokens of a text.}
	\label{ent.ling}
\end{table}
Unlike the other genres, no single feature was remarkably more frequent in credible or non-credible entertainment news (Table \ref{ent.ling}). However, more features overall showed a small to medium effect. A medium effect was observed on the difference between prepositional phrases and type-token ratio, indicating they are more frequent in non-credible news. Second person and demonstrative pronouns were also more frequent in non-credible news, but the difference is much smaller. A moderate positive relationship between first person pronouns and private verbs (0.38) was observed in credible news, suggesting that opinions are more often stated through the first person in credible news. These observations appear consistent with the theory that liars attempt to distance themselves from the lie by avoiding inclusive pronouns, whereas truth-tellers do not \citep{Wawer2023,Addawood2019,Newman2003,Dzindolet2005}.

\paragraph{Health:} 
\label{ling.hel}
\begin{table}[]
	\resizebox{\columnwidth}{!}{
		\begin{tabular}{lccc}
			\hline
			\multicolumn{1}{c}{\textbf{Feature}} & \textbf{Cohen's $d$} & \textbf{Credible Mean} & \textbf{Non-credible Mean} \\ \hline
			Emphatics (EMPH)                     & 0.40               & 0.66                   & 0.18                    \\
			Public Verbs (PUBV)                  & 0.67               & 1.31                   & 0.43                    \\
			Agentless Passives (PASS)            & 0.48               & -0.14                  & 0.33                    \\
			Private Verbs (PRIV)                 & 0.44               & -0.55                  & -0.24                   \\
			Third Person Pronouns (TPP3)         & 0.39               & -0.39                  & -0.06                   \\
			Time Adverbials (TIME)               & 0.39               & -0.55                  & -0.15                   \\
			Split Auxiliaries (SPAU)             & 0.40               & -0.80                  & -0.25                   \\ \hline
		\end{tabular}}
	\caption{Features which show the greatest effect size in \textbf{Health} news. PRIV verbs refer to a mental activity or sensation of which an external observer is not directly aware (e.g., `think' or `feel'). TIME includes temporal adverbs, such as `now' or `shortly'.}
	\label{health.ling} 
\end{table}
The frequency of public verbs was much higher in credible than deceptive health news. Emphatics were more frequent in credible news, which may be surprising considering similar features (i.e. superlatives and intensifiers) have been linked with unreliable news \citep{Mahyoob2020,Francis2018}. Passive constructions and private verbs are more frequent in non-credible health news (Table \ref{health.ling}), which also contradicts \citet{Mahyoob2020}. Several other features, such as split auxiliaries, adverbs of time, and third person pronouns, were mildly more frequent in non-credible text. There is a moderate positive relationship between conditionals and present tense verbs (0.32), present tense verbs and second person pronouns (0.41), and second person pronouns and possibility modals (0.34) in credible health news texts.

Non-credible health news showed a moderate negative relationship between average word length and past tense verbs (-0.41), while there was a stronger positive relationship between average word length and pure nouns (0.48). 

\paragraph{Science:} 
\label{ling.sci}
\begin{table}[]
	\resizebox{\columnwidth}{!}{%
		\begin{tabular}{lccc}
			\hline
			\multicolumn{1}{c}{\textbf{Features}}     & \textbf{Cohen's $d$} & \textbf{Credible Mean} & \textbf{Non-credible Mean} \\ \hline
			Pronoun `it' (PIT)                        & 0.64               & 0.44                   & -0.19                   \\
			Emphatics (EMPH)                          & 0.36               & 0.27                   & -0.27                   \\
			`That' Verb Complements (THVC)            & 0.45               & 0.07                   & 0.55                    \\
			Present Participle Whiz-Deletion (WZPRES) & 0.42               & 0.55                   & 1.39                    \\
			Agentless Passives (PASS)                 & 0.48               & -0.39                  & 0.00                    \\ \hline
		\end{tabular}}
		\caption{Linguistic features which show the greatest disparity in \textbf{Science} news. WZPRES refers to \textit{whiz}-deletion, where a wh-word and `be' are deleted in a relative clause.}
		\label{sci.ling}
\end{table}
Emphatics are more common in credible news, although the effect size is somewhat small. Credible news also contrasted with non-credible news in frequency of the pronoun `it' (Table \ref{sci.ling}). Non-credible science news displayed more passive constructions, \textit{whiz}-deletion, and verb complements with `that'. This is contradictory to \citet{Mahyoob2020}, where passive voice was found to be more frequent in credible news. Demonstratives and demonstrative pronouns have a mild positive relationship between adverbs (0.36 and 0.33 respectively), present tense verbs (0.32), and main verb `be' (0.31) in credible news text. 

Credible news also showed a moderate relationship positive correlation between adverbs and emphatics (0.37), present tense (0.39), and the main verb `be' (0.38). Similar to economy news, adjectives were positively correlated with average word length in the non-credible corpus (0.42). Modifiers, especially adjectives, have also been found to be features of non-credible news in other studies \citep{Addawood2019,Francis2018}.

\paragraph{Sports:}
\label{ling.sports}
\begin{table}[]
	\resizebox{\columnwidth}{!}{
		\begin{tabular}{lccc}
			\hline
			\multicolumn{1}{c}{\textbf{Feature}} & \textbf{Cohen's $d$} & \textbf{Credible Mean} & \textbf{Non-credible Mean} \\ \hline
			Emphatics (EMPH)                     & 0.43               & 0.28                   & -0.23                   \\
			Adjectives (JJ)                      & 0.43               & -0.07                  & -0.46                   \\
			'That' Verb Complements (THVC)       & 0.57               & -0.56                  & 0.13                    \\
			Type-Token Ratio (TTR)               & 0.49               & -0.41                  & -1.40                   \\ \hline
		\end{tabular}}
	\caption{Linguistic features which show the greatest disparity between deceptive and credible news for \textbf{Sports} news.}
	\label{sports.ling} 
\end{table}
There is a medium effect on the difference for type-token ratio, indicating that credible sports writing may be slightly more linguistically diverse (Table \ref{sports.ling}). There was an opposite trend observed in the frequency of emphatics, with credible news reporting a positive mean and deceptive a negative mean. Verb complements with `that' are also much more frequent in non-credible news. Adjectives were less common in non-credible sports news than credible, which contradicts observations for the previous topics and existing literature \citep{Addawood2019,Francis2018}. A mild positive correlation between present tense verbs, demonstrative pronouns (0.34), and first person pronouns (0.31) was observed in non-credible sports news.

\subsection{Discussion}
The higher frequency of past tense verbs, public verbs, and subordinate `that' deletion, combined with the positive correlation between past tense verbs and third person pronouns, implies that the higher D2 score for credible economy news may be from quotations or paraphrasing. Credible content on the economy may reference experts who explain and interpret economic concepts and trends for the reader. The relationship between conditionals and second person pronouns suggests that economic impact on the audience and society may be discussed. In this regard, credible economy news may contradict previous deception research claiming that features like quotations and expressions of possibility reveal uncertainty. 

The relationship between private verbs and \textit{wh}-clauses, along with the low frequency of past tense verbs, non-credible economy news might express uncertainty and employ more appeals to emotion. As mentioned, evocation of emotion and uncertainty are features typically utilised in deceptive language \citep{Igwebuike2021,Newman2003,Wawer2023,Dzindolet2005}.

The correlation between first person pronouns and private verbs implies that credible entertainment news includes more conjecture. The lower score for D4 indicates that author opinion is not overtly expressed in credible entertainment news, so the correlation between these two linguistic features may be due to reporting on rumours. The slightly higher frequency of second person and demonstrative pronouns in non-credible entertainment news may be explained by the presence of sensationalist statements often employed in tabloids.

The frequency of emphatics and public verbs, along with the correlation between present tense verbs, conditionals, and second person pronouns suggests that credible health news may include advice to readers. Furthermore, the positive relationship between second person pronouns and possibility modals suggests that credible news may discuss the effects of health related content on the reader. A general survey of credible health related headlines reveals that content often covers medical advancements and changes in legislation which could potentially impact readers. Deceptive health news showed more passive constructions and private verbs than credible, which appears to oppose \citet{Mahyoob2020}'s findings.

The higher frequency of emphatics and pronoun `it', in addition to the relationship between demonstratives and adverbs, suggests that credible science news enthusiastically discusses concepts and objects more than individuals. 

Non-credible sports news uses more `that' verb complements and public verbs. It also demonstrated a positive relationship between present tense verbs, demonstrative, and first person pronouns. This may hint that non-credible sports news includes more commentary. The use and misuse of quotations has been identified as a feature of deceptive writing that conveys uncertainty \citep{Kasseropoulos2021, Bell1991}. Given the low score for expression of persuasion, comments in the first person may be attributed to quotes from athletes or sports officials.

Overall, notable differences were observed between non-credible and credible news in all topics. Perhaps unsurprisingly, sports news showed the least difference between the corpora. As argued by \citet{Fowler1991}, conversation is a vehicle of ideology. Ideological values are likely more readily expressed through the topics of economy and health rather than sports. Dimension scores for deceptive news indicate a generally higher level of technical language and formality. Although some research has found fake news to be less complex \citep{Kasseropoulos2021,Silva2022}, this is consistent with Fowler's analysis. \citet{Fowler1991} noted that aspects of hysterical style include an excess of negative emotion conveyed through technical jargon, metaphor, and quantification. 

\section{Conclusions and future work}
Although differences between credible and non-credible news were observed in all topics, details varied considerably. Importantly, while not all cues of deception identified in previous literature were present in every domain, most topics showed at least one characteristic of deceptive language. \citet{Plank2016} argues that many NLP models suffer when applied to the real world because they are trained on canonical data. While there appears to be some characteristics of deceptive news text that are shared, primarily technical language, topic differences between credible and non-credible news are too varied for tasks involving canonical `fake news'. For non-credible news classification tasks, it is beneficial to focus on adapting approaches to specific topics.

Appeals to emotion and language that elicits negative emotion have been identified as features of deceptive language and text \citep{Newman2003,Wawer2023,Igwebuike2021}. In the future, it will be useful to investigate negativity in non-credible news topics by using psycholinguistic features with Linguistic Inquiry and Word Count (LIWC). In light of recent technological advancements, it would also be interesting to compare LLM generated non-credible news to see if features of deception are also present in generated news. Further research may also look into stylistic differences in news from other regions as deception cues are likely to vary based on culture and language.

\section{Limitations}
The analysis would benefit from further investigation with discourse processing and the inclusion of psycholinguistic features. While it is not possible to investigate all latent variables that may affect differences between genres and deceptive writing, it would be beneficial to include an analysis on the impact of negation and negativity in non-credible news text. Additionally, as the data is limited to English from North America, it is possible that cultural differences related to deception might result in different patterns. Relatedly, writer demographic (e.g. age, sex, nationality, etc.) may affect deception cues in a text. However, such information is often difficult to discover and may be ethically troublesome to include. The features investigated in this paper are a good focal point, as they have been well studied and are easily accessible.

%Another potential issue lies with labelling data based on the publisher. While this may assume that a publisher always publishes credible or non-credible news, this is probably a reasonable assumption. It is unlikely that a non-credible news publisher will output an article of the same journalistic quality as the New York Times. Furthermore, credible publishers maintain their credible status with rating organisations by issuing timely corrections and avoiding overt bias. As it is unfeasible to manually fact and bias check every article published by an outlet, basing labels on the publisher is a viable alternative.

\section{Ethical concerns}
The decision to consider a piece or source of news media deceptive can be problematic. Relying on simple falsity is often not reliable, as being untruthful is not always to lie \citep{Simpson1992}. Furthermore, labels like `fake news' are often used as political tools to discredit unfavourable interpretations. Even efforts to protect readers from legitimate disinformation can be perceived as censorship. Bias is an inherent part of news, as institutions always report from an angle which is socially, politically, and economically situated \citep{Fowler1991}. 

As a researcher of non-credible news, it is important to consider the implications of attaching labels to media. This is even more important for automatic classification, where false positives and negatives can be particularly damaging. An argument can also be made that exposing characteristics which differentiate credible from deceptive news may assist nefarious actors in creating more convincing fakes. While this is a possibility, it is probably more likely that `fake news' creators are already aware of these differences. Even if research on non-credible news can be exploited, the potential misuse of one's research is not a sufficient argument against the pursuit of knowledge.

\section{Data and code availability}
Many deceptive news sites used in the corpus have become defunct or are no longer updated, but access may be possible through internet archives. To respect copyright, data has not been made public. Code and a description of the data are available on Github. Interested parties are encouraged to reach out to the authors for more information.

%\section{Bibliographical References}
%\label{sec:reference}

% Entries for the entire Anthology, followed by custom entries
\bibliography{lrec_news_text}
\bibliographystyle{acl_natbib}

%\appendix
%
%\section{Example Appendix}
%\label{sec:appendix}
%
%This is a section in the appendix.

\end{document}